\documentclass[11pt,a4paper]{article}
\usepackage[hyperref]{emnlp2020}
\usepackage{times}
\usepackage{url}
\usepackage{algorithm}
\usepackage{algorithmic}
\usepackage{latexsym}
\usepackage{microtype}
\usepackage{graphicx}
\usepackage{subfigure}
\usepackage{booktabs}
\usepackage{amsmath}
\usepackage{amsthm}
\usepackage{enumitem}
\usepackage{color,amsfonts}
\usepackage{multirow}

\newcommand\vx{\pmb{x}}

\newcommand\vxs{\pmb{x}_{S}}

\newcommand\vxscomp{\pmb{x}_{\bar{S}}}
\newcommand\vxsicomp{\pmb{x}_{\bar{S_i}}}

\newcommand\vh{\pmb{h}}

\newcommand\softmax{\textrm{softmax}}

\newcommand\vZ{\mathcal{Z}}

\DeclareMathOperator*{\argmax}{argmax}

\newcommand{\mask}{{\textsc{[Mask]}}}

\usepackage{microtype}

\aclfinalcopy 
\def\aclpaperid{3434} 

\title{Effective Unsupervised Domain Adaptation with \\ Adversarially Trained  Language Models   }

\author{Thuy-Trang Vu$^\diamondsuit$ 
    \And Dinh Phung$^\dagger$ \\
     Faculty of Information Technology,  Monash University, Australia \\
      \texttt{$^\diamondsuit$\{trang.vuthithuy\},$^\dagger$\{first.last\}@monash.edu} \\
    \And Gholamreza Haffari$^\dagger$ 
       }

\date{}

\begin{document}
\maketitle
\begin{abstract}

Recent work has shown the importance of adaptation of  broad-coverage contextualised embedding models on the domain of the target task of interest. 
Current self-supervised adaptation methods are simplistic, as the training signal comes from a small percentage of \emph{randomly} masked-out tokens.
In this paper, we show that careful masking strategies can bridge  the knowledge gap of masked language models (MLMs) about the domains more effectively by allocating  self-supervision where it is  needed.
Furthermore, we propose an effective training strategy by adversarially masking out  those  tokens which are harder to reconstruct by the underlying MLM. %
The adversarial objective leads to a challenging combinatorial optimisation problem over  \emph{subsets} of  tokens, which we tackle efficiently through relaxation to a
variational lower-bound and dynamic programming.
On six unsupervised domain adaptation tasks involving named entity recognition, our method strongly outperforms the random masking strategy and achieves up to +1.64 F1 score improvements.

\end{abstract}

\section{Introduction}
Contextualised word embedding models  are becoming the foundation of state-of-the-art NLP systems~\citep{peters-etal-2018-deep,devlin-etal-2019-bert,roberta,yang2019xlnet,2019t5,brown2020language, Clark2020}.
%
%
%
These models are pretrained on large amounts of raw text using self-supervision to reduce the labeled data requirement of  target tasks of interest by providing useful feature representations~\citep{wang2019glue}. 
Recent work has shown the importance of further training of pre-trained masked language models (MLMs) on the target domain text, as the benefits of their contextualised representations can deteriorate substantially in the presence of domain mismatch~\citep{Ma2019, Xu2019, Wang2019, Gururangan2020}.
This is particularly crucial in unsupervised domain adaptation (UDA), where   there is no labeled data in the target domain \citep{han-eisenstein-2019-unsupervised} and the knowledge from source domain labeled data is transferred to the target domain via a common representation space.
%
However, current  self-supervised adaptation methods are simplistic, as the training signal comes from a small percentage of \emph{randomly} masked-out  tokens.
Thus, it remains to investigate whether there exist more effective  self-supervision strategies to  bridge the knowledge gap of MLMs  about  the  domains to yield higher-quality adapted models. 
A key principle of UDA is to learn a common embedding space of both domains which enables transferring a learned model on source task to target task.
It is typically done by further pretraining the MLM on a combination of both source and target data.
%
Selecting relevant training examples has been shown to be effective in preventing the negative transfer and boosting the performance of adapted models~\citep{moore-lewis-2010-intelligent,ruder-plank-2017-learning}.
Therefore, we hypothesise that the computational effort of the further pretraining should concentrate more on learning words which are specific to the target domain or undergo semantic/syntactic shifts between the domains. 
%

{In this paper, we show that the adapted model can benefit from careful masking strategy and propose an adversarial objective to select subsets for which the current underlying MLM is less confident. This objective raises a challenging combinatorial optimisation problem which we tackle by optimising its variational lower bound. 
We propose a training algorithm which alternates between tightening the variational lower bound and learning the parameters of the underlying MLM.
This involves proposing  an efficient dynamic programming (DP) algorithm to sample from the distribution over the space of masking subsets, and an effective method based on  Gumbel softmax to differentiate through the subset sampling algorithm.
%

We evaluate our adversarial strategy against the random masking and other heuristic strategies including POS-based and uncertainty-based selection on UDA problem of six NER span prediction tasks. These tasks involve adapting NER systems from the news domain to financial, twitter, and biomedical domains. Given the same computational budget for further self-supervising the MLM, the experimental results show that our adversarial approach is more effective than the other approaches, achieving improvements up to +1.64 points in Fscore and +2.23 in token accuracy compared to the random masking strategy.
%
}

\section{Uunsupervised DA with Masked LMs}
\label{sec:uda}


 \paragraph{UDA-MLM.} This paper focuses on the UDA problem where we leverage the labeled data of a related source task to learn a model for a target task without accessing to its labels. We follow the two-step UDA procedure proposed in AdaptaBERT consisting of a domain tuning step to learn a common embedding space for both domains and a task tuning step to learn to predict task labels on source labeled data~\citep{han-eisenstein-2019-unsupervised}. The learned model on the source task can be then zero-shot transferred to the target task thanks to the assumption that these tasks share the same label distribution. 

 This domain-then-task-tuning procedure resembles the pretrain-then-finetuning paradigm of MLM where the domain tuning shares the same training objective with the pretraining. 
 In domain tuning step, off-the-shelf MLM is further pretrained on an equal mixture of randomly masked-out source and target domain data.  
 
\paragraph{Self-Supervision.} The training principle of MLM is based on self-supervised learning where the labels are automatically generated from unlabeled data. The labels are generated by covering some parts of the input, then asking the model to predict them given the rest of the input.

More specifically, a subset of tokens is sampled from the original sequence $\vx$ and replaced with \mask \ or other random tokens~\cite{devlin-etal-2019-bert}.\footnote{In BERT implementation, 15\% tokens in $\vx$ are selected; among them 80\% are replaced with \mask, 10\% are replaced with random tokens, and 10\% are kept unchanged.} 
Without loss of generality, we assume that all sampled tokens are replaced with \mask.
Let us denote the set of masked out indices by $S$, the ground truth tokens by $\vxs = \{x_i | i \in S\}$, 
and the resulting \emph{puzzle} by $\vxscomp$ which is generated by masking out the sentence tokens with indices in $S$. 
The training objective is to minimize the negative log likelihood of the ground truth,
\begin{equation}
\min_\theta -\sum_{\vx \in \mathcal{D}}\log Pr(\vxs|\vxscomp;\mathcal{B}_\theta)
\label{eq:mlm}
\end{equation}
where $\mathcal{B}_\theta$ is 
the MLM parameterised by $\theta$, and $\mathcal{D}$ is the training corpus.

\section{Adversarially Trained Masked LMs }
 
 Given a finite  computational budget, we argue that it should be spent wisely on new tokens or those having semantic/syntactic shifts between the two domains. Our observation is that such tokens would pose more challenging puzzles to the MLM, i.e. the model is less confident when predicting them. Therefore, we propose to strategically select  subsets for which the current underlying MLM $\mathcal{B}_\theta$ is less confident about its predictions:
\begin{equation}
\min_\theta \max_{S \in \mathcal{S}_K} -\log Pr(\vxs|\vxscomp;\mathcal{B}_\theta)
\label{eq:general-obj}
\end{equation}
Henceforth, we assume that the size of the masked set $K$ for a given sentence $\vx$ is fixed.  For example in BERT \cite{devlin-etal-2019-bert}, $K$ is taken to be $15\%\times|\vx|$ where $|\vx|$ denotes the length of the sentence.
We denote all possible subsets of  indices in a sentence with a fixed size by $\mathcal{S}_K$.



\subsection{Our Variational Formulation}
\label{sec:adversarial}
\begin{figure*}
    \centerline{\includegraphics[scale=0.63]{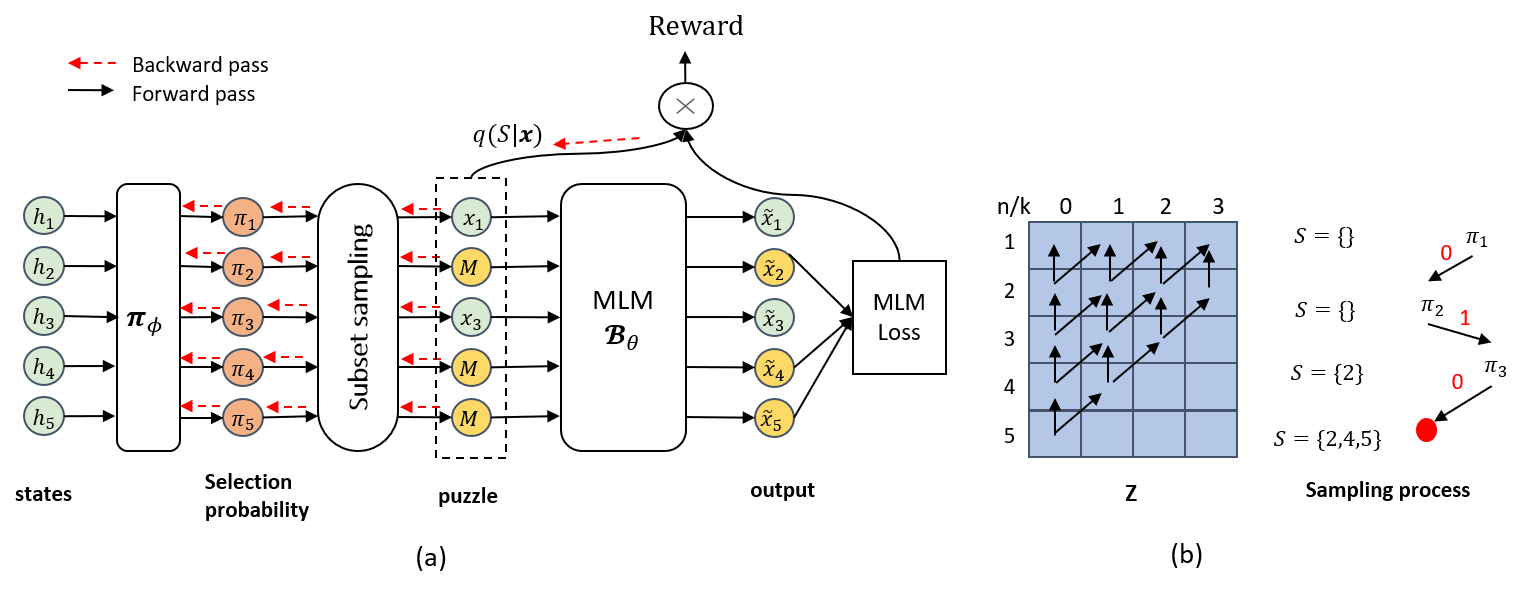}}
    \caption{(a) Our adversarial learned masking strategy for MLM includes a puzzle generator to estimate selection probability, a subset sampling procedure and the MLM model. The red dash arrow shows the gradient flow when updating the puzzle generator. (b) Masked subset sampling procedure with dynamic programming.}
    \label{fig:adv-mlm-model}
\end{figure*}

The masking strategy learning problem described in eqn~(\ref{eq:general-obj}) is a minimax game of two players: the puzzle generator to select the  subset  resulting in the most challenging puzzle, and the MLM $\mathcal{B}_\theta$ to best solve the puzzle by reconstructing the masked tokens correctly. 
As optimising over the subsets is a hard combinatorial problem over the discrete space of $\mathcal{S}_K$, we are going to convert it to a continuous optimisation problem.

We establish a variational lower bound of the objective function over $S$ using the following inequality,
%
\begin{eqnarray}
\max_{S \in \mathcal{S}_K} -\log Pr(\vxs|\vxscomp;\mathcal{B}_\theta) \geq \\ \max_{\phi} \sum_{S \in \mathcal{S}_K} - q(S|\vx;\pi_\phi) \log Pr(\vxs|\vxscomp;\mathcal{B}_\theta)
\end{eqnarray}
where $q(.)$ is the variational distribution provided by a neural network $\pi_{\phi}$. This variational distribution $q(S|\vx;\pi_\phi)$ estimates the distribution over all subset of size $K$. It is straightforward to see that the weighted sum of negative log likelihood of all possible subsets is always less than the max value of them.
%
Our minimax training objective is thus,
\begin{eqnarray}
\label{eqn:minimax} \min_{\theta} \max_{\phi} \sum_{S \in \mathcal{S}_K} -q(S|\vx;\pi_\phi) \log Pr(\vxs|\vxscomp;\mathcal{B}_\theta) \\
\label{eqn:var} q(S|\vx,\pi_{\phi}) = \displaystyle \prod_{i \in S} \pi_{\phi}(i|\vx)\prod_{i'\not\in S} (1-\pi_{\phi}(i'|\vx))/\mathcal{Z}
\end{eqnarray}
where  $\mathcal{Z}$ is the partition function making sure the probability distribution sums to one,
\begin{equation}
\vZ = \sum_{S' \in \mathcal{S}_K} \prod_{i \in S'} \pi_{\phi}(i|\vx)\prod_{i' \notin S'} (1-\pi(i'|\vx)).
\label{eqn:partition}
\end{equation}
The number of possible subsets  is $|\mathcal{S}_K|=\binom{|\vx|}{K}$, which grows exponentially with respect to $K$. 
In \S \ref{sec:subsets}, we provide efficient dynamic programming algorithm for  computing the partition function and sampling from this exponentially large combinatorial space.
In the following, we  present our model architecture and training algorithm for the puzzle generator $\phi$ and MLM $\theta$ parameters based on the variational training objective in eqn~(\ref{eqn:minimax}).

\subsection{Model Architecture}
We learn the masking strategy through the puzzle generator network as shown in Figure~\ref{fig:adv-mlm-model}. It is a feed-forward neural network assigning a selection probability $\pi_\phi(i|\vx)$ for each index $i$ given the original sentence $\vx$, where $\phi$ denote the parameters. Inputs to the puzzle generator are the feature representations $\{\vh_i\}_{i=1}^n$ of the original sequence $\{\vx_i\}_{i=1}^n$. More specifically, they are output of the last hidden states of the MLM. The probability of perform masking at position $i$ is computed by applying sigmoid function over the feed-forward net output 
$\pi_{\phi}(i|\vx) = \sigma(\textrm{FFNN}(\vh_i))$. From these probabilities, we can sample the masked positions in order to further train the underlying MLM $\mathcal{B}_\theta$.

\subsection{Optimising the Variational Bound}
We use an alternating optimisation algorithm to train the MLM $\mathcal{B}_{\theta}$ and the puzzle generator $\pi_{\phi}$ (Algorithm~\ref{alg:training-procedure}). The update frequency for $\pi_{\phi}$ is determined via a mixing hyperparameter $\beta$.

\paragraph{Training the MLM.} Fixing the puzzle generator, we can train the underlying MLM model using gradient descent on MLM objective in eqn~(\ref{eq:mlm}),
\begin{equation}
\min_\theta \mathbb{E}_{q(S|\vx;\pi_\phi)}[-\log Pr(\vxs|\vxscomp;\mathcal{B}_\theta)]
\label{eq:update_theta}
\end{equation}
where we approximate the expectation by sampling. That is, $\mathbb{E}_{q(S|\vx;\pi_\phi)}[-\log Pr(\vxs|\vxscomp;\mathcal{B}_\theta)] $ is approximated by 
\begin{equation}
\frac{1}{M} \sum_{m=1}^M -\log Pr(\vx_{S_m}|\vx_{\bar{S}_m};\mathcal{B}_\theta)\
\end{equation}
where $S_m \sim q(S|\vx;\pi_\phi)$. In \S \ref{sec:sampling}, we present an efficient sampling algorithm based on a sequential decision making process involving discrete choices, i.e. whether to include an index $i$ or not.

\paragraph{Training the Puzzle Generator.} Fixing the MLM, we can train the puzzle generator by considering $-\log Pr(\vx_S|\vx_{\bar{S}}; \mathcal{B}_{\theta})$ as the reward, and aim to optimise the expected reward,
\begin{eqnarray}
\max_{\phi}\mathbb{E}_{q(S|\vx;\pi_\phi)}[-\log Pr(\vxs|\vxscomp;\mathcal{B}_\theta)].
\label{eq:update-phi}
\end{eqnarray} 
We may aim to sample multiple index sets $\{S_1,..,S_M\}$ from $q(S|\vx;\pi_{\phi})$, and then optimise the parameters of the puzzle generator by maximizing the Monte Carlo estimate of the expected reward. 
However, as sampling each index set $S_m$ corresponds to a sequential decision making process involving \emph{discrete} choices, we cannot backpropagate through the sampling process to learn the parameters of the puzzle generator network. Therefore, we rely on the Gumbel-Softmax trick~\citep{gumbel} to deal with this issue and backpropagate through the parameters of $\pi_\phi$, which we will cover in \S \ref{sec:diff}. 

\begin{algorithm}[tb]
\textbf{Input}: data $\mathcal{D}$, update freq. $\beta$,
                masking size $K$\\
\textbf{Output}: generator $\pi_\phi$, MLM $\mathcal{B}_\theta$\\
\begin{algorithmic}[1]
\STATE Let $\phi \gets \phi_0; \theta \gets \theta_0$
\WHILE{stopping condition is not met}
\FOR {$\vx \in \mathcal{D}$} 
    \STATE $S, q(S) \gets \textrm{subsetSampling}(\vx,\pi_\phi,K) $
    \STATE Update the MLM using Eq.~(\ref{eq:update_theta})
    \IF{coinToss($\beta$)==Head}    
    \STATE Compute reward \\
    $r \gets -\log Pr(\vx_{S}|\vx_{\bar{S}};\mathcal{B}_\theta)$
    \STATE Update the generator using Eq.~(\ref{eq:update-phi})
    \ENDIF
\ENDFOR
\ENDWHILE
\STATE \textbf{return} $\theta, \phi$
\end{algorithmic}
\caption{Adversarial Training Procedure}
\label{alg:training-procedure}
\end{algorithm}


\section{Sampling and Differentiating  Subsets}
\label{sec:subsets} 
\subsection{A DP for the Partition Function} 
%

In order to sample from the variational distribution in eqn (\ref{eqn:var}), we need to compute its partition function in eqn~(\ref{eqn:partition}). 
Interestingly, the partition function  can be computed using dynamic programming (DP).   

Let us denote by $Z(j,k)$ the partition function of all subsets of  size $k$ from the index set $\{j,..,|\vx|\}$. Hence, the partition function of the $q$ distribution is  $Z(1,K)$. The DP relationship can be written as,
\begin{align}
\begin{split}
    Z(j-1,k) &=(1-\pi(j-1|\vx))Z(j,k) \\ 
    &+ \pi_{\phi}(j-1|\vx) Z(j,k-1).
\end{split}
\label{eq:dp}
\end{align}
The initial conditions are $Z(j,0) = 1$ and 
\begin{equation}
Z(|\vx|-k+1,k) = \prod_{i=|\vx|-k+1}^{|\vx|}  \pi_{\phi}(j|\vx)
\end{equation}
corresponding to two special terminal cases in selection process in which we have picked all $K$ indices, and we need to select all indices left to fulfil $K$. 

This amounts to a DP algorithm with the time complexity $\mathcal{O}(K|\vx|)$.

\begin{algorithm}[tb]
\textbf{Function}: subsetSampling \\
\textbf{Input}: datapoint $\vx$, prob. $\pi_\phi$,
                masking size $K$\\
\textbf{Output}: subset $S$, sample $\log$ probability $l$ \\
 \begin{algorithmic}[1]
\STATE Let $S \gets \emptyset$; $l \gets 0$; $j \gets 0$
\STATE Calculate DP table $Z$ using Eq.~(\ref{eq:dp})
\WHILE{$|S| < K$}
    \STATE $j \gets j + 1$
    \STATE $q_{j,Y} \leftarrow q_j(\textrm{Y}|S_{j-1}, \pi_{\phi})$ // using eqn~(\ref{eqn:stochastic:policy})
    \STATE $q_{j,N} \leftarrow 1 - q_{j,Y}$ 
    \STATE $\epsilon_{j,Y}, \epsilon_{j,N} \sim \textrm{Gumbel}(0,1) $
    \STATE $o_j \gets \argmax_{o \in \{Y,N\}}\log q_{j,o} + \epsilon_{j,o}$
    
     \STATE $l \mathrel{+}=  \log \softmax(\log q_{j,o} + \epsilon_{j,o})\big\rvert_{o=o_j}$
    \IF {$o_j == Y$}
        \STATE $S \gets S \cup \{j\}$
    \ENDIF
\ENDWHILE
\STATE \textbf{return}   $S,l$
\end{algorithmic}
\caption{Sampling Procedure}
\label{alg:sample-student}
\end{algorithm}

\begin{table*}[h]
\begin{center}
\scalebox{0.93}{
\begin{tabular}{l||lccc|lc}
\toprule
&\multicolumn{1}{c}{\textbf{NER}} & \multicolumn{3}{c}{\textbf{Num. tokens/Num. sent.}} & \multicolumn{1}{|c}{\textbf{Unlab.}} &\multicolumn{1}{c}{\textbf{Num. tokens}}\\
\multicolumn{1}{c||}{\textbf{Domain}} & \multicolumn{1}{c}{\textbf{Dataset}} & \multicolumn{1}{c}{\textbf{Train}} & \multicolumn{1}{c}{\textbf{Dev.}} & \multicolumn{1}{c}{\textbf{Test}} & \multicolumn{1}{|c}{\textbf{Corpus}} & \multicolumn{1}{c}{\textbf{/Num. sent.}} \\
\midrule
\textsc{News} & CoNLL2003 & 203.6k/14.0k & 51.36k/3.3k & 46.4k/3.5k & - & - \\
\textsc{Tweet} & WNUT2016 & 46.5k/2.4k & 16.3K/1k & 61.9k/3.8k & Sentiment140 & 20M/1.4M\\
\textsc{Fin} & FIN & 41.0k/1.2k & - & 13.3k/303 & SEC Filing 2019 & 155M/5.5M \\
\textsc{BioMed}& JNLPBA  & 445.0k/16.8k & 47.5k/1.7k & 101.0k/3.9k   & PubMed & 4.3B/181M \\
\textsc{BioMed}  & BC2GM  & 355.4k/12.6k & 71.0k/2.5k & 143.5k/5.0k  & PubMed & 4.3B/181M \\
\textsc{BioMed}& BioNLP09 & 227.7k/7.5k & 44.2k/1.4k & 74.6k/2.5k   & PubMed & 4.3B/181M\\   
\textsc{BioMed}& BioNLP11EPI & 161.6k/5.7k & 54.8k/1.9k & 116.1k/4.1k   & PubMed & 4.3B/181M\\
\bottomrule                 
\end{tabular}
}
\caption{Data statistics of named entity span prediction tasks and unlabled additional pretraining corpus.}
\label{tab:dataset}
\end{center}
\end{table*}
\subsection{Subset Sampling for MLMs}
\label{sec:sampling}

The DP in the previous section also gives rise to the sampling procedure. Given a partial random subset $S_{j-1}$ with elements chosen from the indices $\{1,..,j-1\}$,  the probability of including the next index $j$, denoted by $q_{j}(\textrm{yes}|S_{j-1},\pi_{\phi})$, is
\begin{equation}
\label{eqn:stochastic:policy}
\frac{\pi_{\phi}(j|\vx)Z(j+1,K-1-|S_{j-1}|)}{Z(j,K-|S_{j-1}|)}
\end{equation}
where $Z(j,k)$ values come from the DP table. Hence, the probability of \emph{not} including the index $j$ is 
\begin{equation}
    q_{j}(\textrm{no}|S_{j-1},\pi_{\phi}) = 1 - q_{j}(\textrm{yes}|S_{j-1},\pi_{\phi}).
\end{equation}
%
In case the next index is chosen to be  in the sample, then $S_{j+1} = S_j \cup \{j+1\}$; otherwise $S_{j+1}=S_j$.

The sampling process entails a sequence of binary decisions (Figure~\ref{fig:adv-mlm-model}.b) in an underlying Markov Decision Process (MDP). 
It is an iterative process, which starts by considering the index one.   At each decision point $j$, the sampler's action space is to whether include (or not include) the index $j$  into the partial sample $S_j$ based on eqn~(\ref{eqn:stochastic:policy}). 
We terminate this process when the partially selected subset has $K$ elements.

The sampling procedure is described in Algorithm~\ref{alg:sample-student}. In our MDP, we actually sample an index by generating Gumbel noise in each  stage, and then select the choice (yes/no) with the maximum probability. 
This enables differentiation through the sampled subset, covered in the next section. 


\subsection{Differentiating via Gumbel-Softmax} 
\label{sec:diff}
Once the sampling process is terminated, we then need to backpropagate through the parameters of $\pi_\phi$, when updating the parameters of the puzzle generator according to eqn (\ref{eq:update-phi}). 

More concretely, let us assume that we would like to sample a subset $S$. As mentioned in previous section, we need to decide about the inclusion of the next index $j$ given the partial sample so far $S_{j-1}$ based on the eqn~(\ref{eqn:stochastic:policy}). Instead of uniform sampling, we can equivalently choose one of these two outcomes as follows
\begin{equation}
    o^*_j =\displaystyle \argmax_{o_j \in \{\textrm{yes,no}\}} \log q_j(o_j|S_{j-1},\pi_\phi)+\epsilon_{o_j}
\end{equation}
where the random noise $\epsilon_{o_j}$ is distributed according to standard Gumbel distribution. Sampling a  subset then amounts to a sequence of $\argmax$ operations. 
To backpropagate through the sampling process, we replace the $\argmax$ operators with $\softmax$, as $\argmax$ is not differentiable. 
That is,
\begin{equation}
\resizebox{0.86\hsize}{!}{$ Pr(o_j) = \frac{\exp(\log q_j(o_j|S_{j-1},\pi_\phi)+\epsilon_{o_j})}{\sum_{o_j'}\exp(\log q_j(o_j'|S_{j-1},\pi_\phi)+\epsilon_{o_j'})}.$}
\end{equation}
%
The $\log$ product of the above probabilities for the  decisions  in a  sampling \emph{path} is returned as  $l$ in Algorithm \ref{alg:sample-student}, which is then used for backpropagation.

\section{Experiments}
We evaluate our proposed masking strategy in UDA for named entity span prediction tasks coming from three different domains.

\subsection{Unsupervised Domain Adaptation Tasks}
\paragraph{Source and Target Domain Tasks.} Our evaluation is focused on the problem of identifying named entity spans in domain-specific text without access to labeled data. The evaluation tasks comes from several named entity recognition (NER) dataset including WNUT2016~\citep{strauss-etal-2016-results}, FIN~\citep{salinas-alvarado-etal-2015-domain}, JNLPBA~\citep{collier-kim-2004-introduction}, BC2GM~\citep{smith2008overview}, BioNLP09~\citep{kim-etal-2009-overview}, and BioNLP11EPI~\citep{kim-etal-2011-overview}. 
Table~\ref{tab:dataset} reports data statistics.
%

 These datasets cover three domains social media (\textsc{Tweets}), financial (\textsc{Fin}) and biomedical (\textsc{BioMed}). We utilize the CoNLL-2003 English NER dataset in newstext domain (\textsc{News}) as the source task and  others as the target. 
 We perform  domain-tuning and source task-tuning, followed by zero-shot transfer  to the target tasks, as described in~\S\ref{sec:uda}. 
 Crucially, we do not use the labels of the training sets of the target tasks, and only use their sentences for domain adaptation.
 Since the number of entity types are different in each task, we convert all the labels to entity span in IBO scheme. This ensures that all tasks share the same set of labels consisting of three tags: I, B, and O.


\paragraph{Extra  Target Domain Unlabeled Corpora.} As the domain tuning step can further benefit from additional unlabeled data, we create target domain  unlabeled datasets from the available corpora of relevant domains. More specifically, we use publicly available corpora, Sentiment140~\citep{go2009twitter}, SEC Filing 2019\footnote{\url{http://people.ischool.berkeley.edu/~khanna/fin10-K/}}~\citep{desolafinbert}  PubMed~\citep{lee2020biobert} for the \textsc{Tweet}, \textsc{Fin} and \textsc{BioMed} domains respectively (Table \ref{tab:dataset}). From the unlabeled corpora, the top 500K and 1M similar sentences to the training set of each target task are extracted based on the average $n$-gram similarity where $1\leq n \leq 4$, resulting in extra target domain unlabeled corpora.

\begin{table*}[!h]
\begin{center}
\scalebox{0.82}{
\begin{tabular}{l||rrrr|rrrr|rrrr}
\toprule
 & \multicolumn{4}{c|}{\textbf{UDA}} & \multicolumn{4}{c}{\textbf{UDA + 500K target-domain}}  & \multicolumn{4}{|c}{\textbf{UDA + 1M target-domain}}\\
\multicolumn{1}{c||}{Task} & 
\multicolumn{1}{c}{rand} & \multicolumn{1}{c}{pos} & \multicolumn{1}{c}{ent} & \multicolumn{1}{c|}{adv} & \multicolumn{1}{c}{rand} & \multicolumn{1}{c}{pos} & \multicolumn{1}{c}{ent} & \multicolumn{1}{c}{adv} &
\multicolumn{1}{|c}{rand} & \multicolumn{1}{c}{pos} & \multicolumn{1}{c}{ent} & \multicolumn{1}{c}{adv}\\
\midrule
WNUT2016 &	 \textbf{47.11} & 46.79$^\dagger$ & 46.95$^\dagger$ & 47.03$^\dagger$ & 46.93 & 47.69$^\dagger$ & 47.84$^\dagger$ & \textbf{48.01}$^\dagger$ & 52.36 &	52.01$^\dagger$ & \textbf{52.74}$^\dagger$ & 52.53$^\dagger$ \\
FIN  & 21.55 & 22.53$^\dagger$ & 22.73$^\dagger$ & \textbf{23.38}$^\dagger$ & 24.70 & 26.70$^\dagger$ & 26.63$^\dagger$ & \textbf{26.85}$^\dagger$ & 25.96 &	26.95$^\dagger$ &	26.96$^\dagger$ & \textbf{28.94}$^\dagger$ \\
JNLPBA  & 27.44 & 28.06$^\dagger$ & 28.22$^\dagger$ & \textbf{30.06}$^\dagger$ & 29.92 & \textbf{30.56}$^\dagger$ & 30.47$^\dagger$ & 30.31$^\dagger$ & 31.01 &	30.91$^\dagger$ &	\textbf{31.59}$^\dagger$ & 31.54$^\dagger$ \\
BC2GM  & 28.31 & 28.50 & \textbf{30.81}$^\dagger$ & 29.01$^\dagger$ & 31.13 & 31.85$^\dagger$ & 31.83$^\dagger$ & \textbf{32.38}$^\dagger$ & 31.35 &	31.70$^\dagger$ &	32.01$^\dagger$ & \textbf{32.49}$^\dagger$ \\
BioNLP09  & 26.37 & 27.53$^\dagger$ & 29.21$^\dagger$ & \textbf{29.24}$^\dagger$ & 31.38 & 31.03$^\dagger$ & 34.33$^\dagger$ & \textbf{35.05}$^\dagger$ & 32.16 &	33.51$^\dagger$ & 34.99$^\dagger$ & \textbf{35.41}$^\dagger$ \\
BioNLP11EPI & 32.69 & 33.51$^\dagger$ & \textbf{34.81}$^\dagger$ & 34.59$^\dagger$ & 42.41 & 42.81$^\dagger$ & \textbf{42.83}$^\dagger$ & 42.64 & 43.11 &	43.47$^\dagger$ & 43.31 & \textbf{43.61}$^\dagger$\\
\midrule
$\Bar{\Delta}$      & - & +0.58 & +1.54 & \textbf{+1.64} & - & +0.70 & +1.26 & \textbf{+1.46} &  & +0.43 & +0.94 & \textbf{+1.43} \\ 
\bottomrule
\end{tabular}
}
\caption{F1 score of name entity span prediction tasks in three UDA scenarios which differ in the amount of additional target-domain data. rand, pos, ent and adv denote the random, POS-based, uncertainty-based, and adversarial masking strategy respectively. $\Bar{\Delta}$ row reports the average improvement over random masking across all tasks. \textbf{Bold} shows the highest score of task on each UDA setting. $^\dagger$~indicates statistically significant difference to the random baseline with p-value $\leq$ 0.05 using bootstrap test.}
\label{tab:uda}
\end{center}
\end{table*}

\subsection{Masking Strategies for MLM Training}
\label{exp:strategy}
We compare our adversarial learned masking strategy approach against random and various heuristic masking strategies which we propose:
\begin{itemize}[leftmargin=*]
  \item \textbf{Random.} Masked tokens are sampled uniformly at random, which  is the common strategy  in the literature~\citep{devlin-etal-2019-bert,roberta}.
  \item \textbf{POS-based strategy.} Masked tokens are sampled according to a non-uniform distribution, where a token's probability depends on its POS tag. The POS tags are obtained using spaCy.\footnote{\url{https://spacy.io/}}
Content tokens such as verb (VERB), noun (N), adjective (ADJ), pronoun (PRON) and adverb
(ADV) tags are assigned higher probability (80\%) than other content-free tokens such as PREP, DET, PUNC (20\%). 
\item \textbf{Uncertainty-based strategy.} We select those tokens for which the current MLM is most uncertain for the reconstruction, where the uncertainty is measured by the entropy. 
That is, we aim to select those tokens with high $\textrm{Entropy}[Pr_{i}(.|\vxsicomp;\mathcal{B}_\theta)]$, where $\vxsicomp$ is the sentence $\vx$ with the $i$th token  masked out, and $Pr_{i}(.|\vxsicomp;\mathcal{B}_\theta)$ is the predictive distribution for the $i$th position in the sentence. 

Calculating the predictive distribution for each position requires one pass through the  network. Hence, it is expensive to use the exact entropy, as it requires $|\vx|$ passes.  We mitigate this cost by using $Pr_i(.|\vx;\mathcal{B}_\theta)$ instead, which conditions on the original unmasked sentence. This estimation only costs one pass through the MLM.



    \item \textbf{Adversarial learned strategy.} 
    The masking strategy is learned adversarially as  in~\S3. The puzzle-generator update frequency $\beta$ (Algorithm~\ref{alg:training-procedure}) is set to 0.3 for all experiments.

\end{itemize} 
These strategies only differ in how we choose the candidate tokens. The number
of to-be-masked tokens is the same in all strategies (15\%). Among them, 80\%
are replaced with $\textrm{[MASK]}$, 10\% are replaced with random words, the
rest are kept unchanged as in \cite{devlin-etal-2019-bert}. 
In our experiments, the masked sentences are generated dynamically on-the-fly.

To evaluate the models, we compute precision, recall and F1 scores on a per token basis. We report average performance of five runs.


\subsection{Implementation Details}
Our implementation is based on Tensorflow library~\citep{abadi2016tensorflow}\footnote{Source code is available at \url{https://github.com/trangvu/mlm4uda}}. We use BERT-Base model architecture which consists of 12 Transformer layers with 12 attention heads and hidden size 768 \citep{devlin-etal-2019-bert} in all our experiments.  We use the cased wordpiece vocabulary provided in the pretrained English model. We set learning rate to 5e-5 for both further pretraining and task tuning. Puzzle generator is a two layer feed-forward network with hidden size 256 and dropout rate 0.1. 

\subsection{Empirical Results}
Under the same computation budget to update the MLM, we evaluate the effect of masking strategy in the domain tuning step under various size of additional target-domain data: none, 500K and 1M. We continue pretraining BERT on a combination of unlabeled source (CoNLL2003), unlabeled target task training data and additional unlabeled target domain data (if any). If target task data is smaller, we oversample it to have equal size to the source data. The model is trained with batch size 32 and max sequence length 128 for 50K steps in 1M target-domain data and 25K steps in other cases. It equals to 3-5 epochs over the training set. After domain tuning, we finetune  the adapted MLM  on the source task labeled training data (CoNLL2003) for three epochs with batch size 32. Finally, we evaluate the resulting model on target task. On the largest dataset, random and POS strategy took around 4 hours on one NVIDIA V100 GPU while entropy and adversarial approach took 5 and 7 hours respectively. The task tuning took about 30 minutes.

\begin{table}[t]
\begin{center}
\scalebox{0.8}{\begin{tabular}{l|l||rrrr}
\toprule  
& Task & rand & mix-pos & mix-ent & mix-adv \\
\midrule
\multirow{6}{*}{\rotatebox[origin=c]{90}{\parbox[c]{3cm}{\centering \textbf{UDA + 500K}}}} 
& WNUT2016 & 46.93& 51.17 & 52.40 & \textbf{52.56} \\ 
& FIN.     & 24.70& 26.95 & 27.36 & \textbf{28.30}   \\ 
& JNLPBA   & 29.92 & 29.22 & 31.65 & \textbf{32.99} \\ 
& BC2GM    & 31.13 &  32.11 & \textbf{32.68} & 32.60 \\ 
& BioNLP09 & 31.38 & 33.17 & 34.27 & \textbf{34.91}  \\ 
& BioNLP11EPI & 42.41 & 42.73 & \textbf{43.43} & 43.08  \\ 
\midrule
&$\Bar{\Delta}$  & - & +3.10& +4.17& \textbf{+4.61}  \\
\midrule
\midrule
\multirow{6}{*}{\rotatebox[origin=c]{90}{\parbox[c]{3cm}{\centering \textbf{UDA + 1M}}}}
& WNUT2016 & 52.36 & 52.40 & 52.64 & \textbf{52.95} \\ 
& FIN.     & 25.96 & 27.86 & 28.51 & \textbf{29.08} \\ 
& JNLPBA   & 31.01 & 31.77 & 32.07 & \textbf{32.26}\\ 
& BC2GM    & 31.35 & 31.76 & 32.43 & \textbf{32.52}\\ 
& BioNLP09 & 32.61 & 34.49 & 35.67 & \textbf{35.78}\\ 
& BioNLP11EPI & 43.11 & 43.96 & \textbf{44.81} & 44.27  \\ 
\midrule
&$\Bar{\Delta}$  & - & +1.05& +1.70& \textbf{+1.82} \\
\bottomrule
\end{tabular}}
\caption{F1 score  in UDA with additional data under several mixed masking strategies. 
\textbf{Bold} shows the highest score of task on each UDA setting.
}
\label{tab:mix-strategy}
\end{center}
\end{table}

Results are shown in Table~\ref{tab:uda}. Overall, strategically masking consistently outperforms random masking in most of the adaptation scenarios and target tasks.  As expected, expanding training data with additional target domain data further improves performance of all models. Comparing to random masking, prioritising content tokens over content-free ones can improve up to 0.7 F1 score in average. By taking the current MLM into account, uncertainty-based selection and adversarial learned strategy boost the score up to 1.64. Our proposed adversarial approach yields highest score in 11 out of 18 cases, and results in the largest improvement over random masking across all tasks in both UDA with and without additional target domain data. 

We further explore the mix of random masking and other masking strategies. 
We hypothesise that the combination strategies can balance the learning of challenging tokens and effortless tokens when forming the common semantic space, hence improve the task performance.
In a minibatch, 50\% of sentences are masked according to the corresponding strategy while the rest are masked randomly. Results are shown in Table~\ref{tab:mix-strategy}. We observe an additional performance to the corresponding single-strategy model across all tasks.



\subsection{Analysis}
\begin{figure}[t]
    \centerline{\includegraphics[scale=0.5]{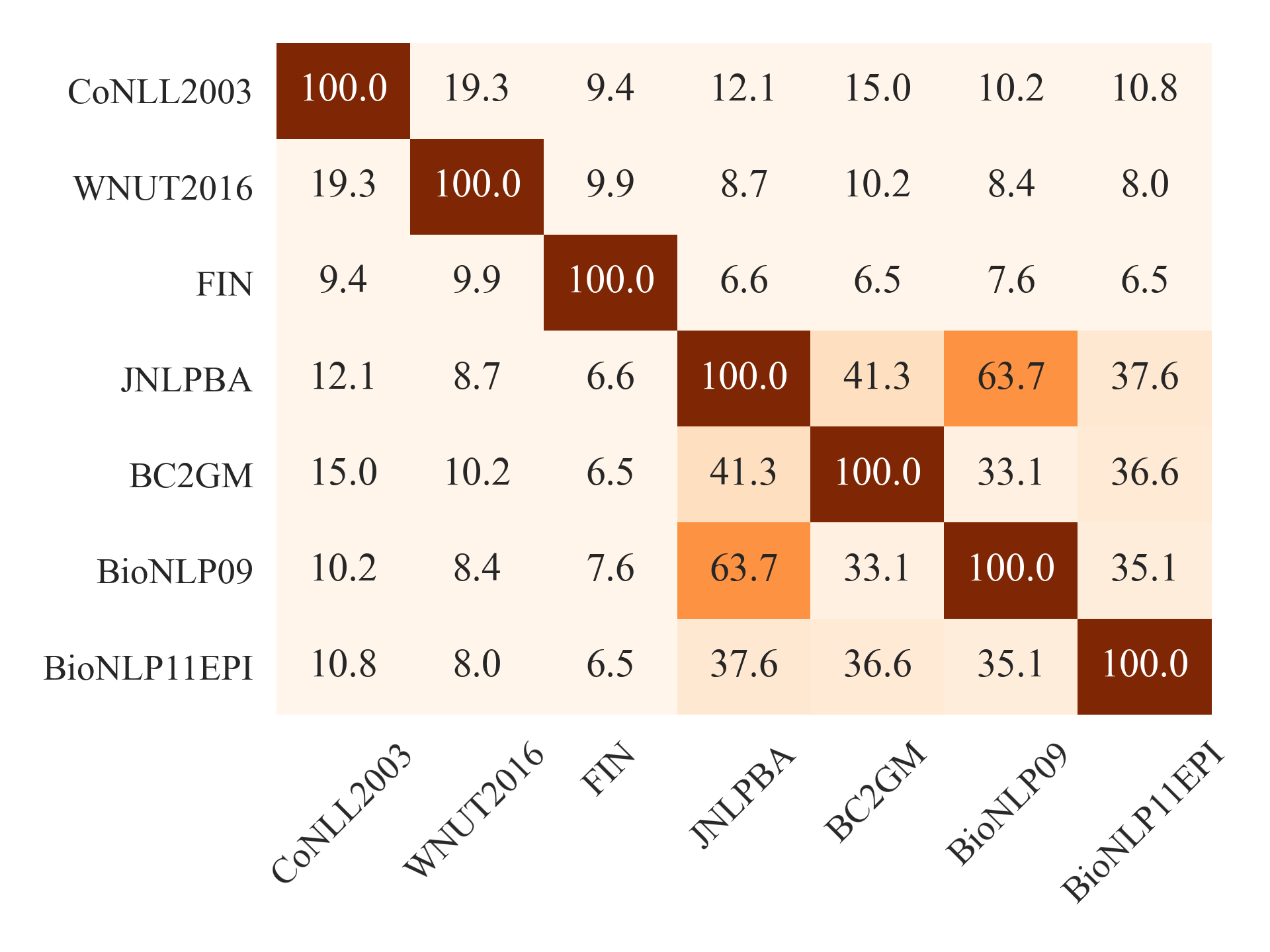}}
    \caption{Vocabulary overlap (\%) between NER tasks. }
    \label{fig:vocab-overlap}
\end{figure}
\paragraph{Domain Similarity.}
We quantify the similarity between source (CoNLL2003) and target domains by vocabulary overlap between the domains (excluding stopwords). Figure~\ref{fig:vocab-overlap}  shows the vocabulary overlap across tasks. As seen, all the target domains are dissimilar to the source domain, with FIN having the lowest overlap. FIN has gained the largest improvement from the adversarial strategy in the UDA results in Tables \ref{tab:uda} and \ref{tab:mix-strategy}.  
As expected, the biomedical datasets have relatively higher vocabulary overlap with each other. 

\begin{figure}[t]
    \centerline{\includegraphics[scale=0.5]{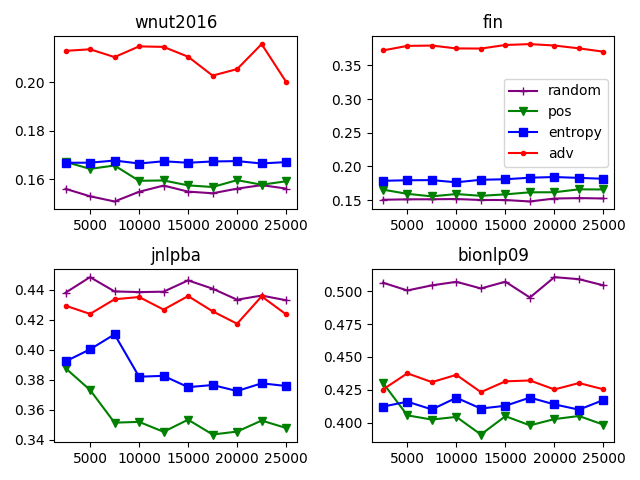}}
    \caption{Average density ratio of masked-out tokens of every 2500 training steps in UDA setting.}
    \label{fig:lexical_ratio}
\end{figure}

\paragraph{Density Ratio of Masked Subsets.}
We analyze the density ratio of masked-out tokens in the target and source domains 
$$r(w) = \max(1 - \frac{Pr_s(w)}{Pr_t(w)},0)$$
where $Pr_s(w)$ and $Pr_t(w)$ is the probability of token $w$ in source and target domains, respectively. These probabilities are according to  unigram language models trained on the training sets of the source and target tasks. The higher value of $r(w)$ means the token $w$ is new or appears more often in the target text than in the source. Figure~\ref{fig:lexical_ratio} plots the density ratio of masked-out tokens during domain tuning time for four UDA tasks. Comparing to other strategies, we observed that adversarial approach tends to select tokens which have higher density ratio, i.e. more significant in the target.

\begin{table}[t]
\begin{center}\begin{tabular}{l|rrrr}
\toprule  
POS Tag & rand & pos & ent & adv \\
\midrule
ADJ & 9\% & 17\% & 11\% & 13\% \\
VERB & 8\% & 16\% & 10\% & 17\%\\
NOUN & 25\% & 51\% & 31\% & 34\% \\
PRON & 1\% & 2\% & 1\% & 3\% \\
ADV & 2\% & 4\% & 2\% & 4\% \\
Others & 55\% & 10\% & 45\% & 29\% \\
\bottomrule
\end{tabular}
\caption{The tag ratio of the POS tags of tokens in masked subset on BIONLP11 under different masking strategies.}
\label{tab:pos_ratio}
\end{center}
\end{table}

\begin{table}[t]
\begin{center}
\scalebox{0.89}{
\begin{tabular}{lrrrr}
\toprule
Task &Model & acc.  & non-OOV & OOV\\
\midrule
 & rand & 23.04 & 21.88  & 24.99 \\
WNUT2016 & pos &23.78 & 22.77 & 25.48    \\
 & ent & 23.95 & \textbf{22.95} & 25.62 \\
 & adv & \textbf{24.20} & 22.79 & \textbf{26.57} \\
\midrule 
 & rand & 27.66 & 25.01 &30.88\\
FIN & pos & 28.51 & 27.23 & 29.36 \\
 & ent & 28.09 & \textbf{27.67} & 31.21\\ 
 & adv & \textbf{29.36} & 27.56 & \textbf{33.90}\\
\midrule
& rand & 7.77 & 7.86  & 7.50 \\
JNLPBA & pos & \textbf{9.74} & \textbf{9.83}  & \textbf{9.50}\\
 & ent & 8.79 & 8.81 & 8.74\\
 & adv & 7.92 & 7.89 & 8.01\\
\midrule
 & rand & 11.38 & 11.35 & 11.48 \\
BC2GM & pos & 13.09 & 12.88 & 13.89\\
 & ent & 13.19 & 12.89 & 14.28 \\
 & adv & \textbf{14.53} & \textbf{14.44} & \textbf{14.84}\\
\midrule
 & rand & 9.49 & 8.88 & 10.2\\
BioNLP09 & pos & 9.45 & 10.51 & 8.22\\
 & ent & 13.11 & 15.67 & 10.14 \\
 & adv & \textbf{14.82} & \textbf{18.45} & \textbf{10.61} \\
 \midrule
 & rand & 13.16 & 27.40 & 6.57 \\
BioNLP11EPI& pos & 14.02 & 28.28 & 7.43 \\
& ent & \textbf{14.28} & \textbf{28.70} & \textbf{7.59} \\
& adv & 13.76 & 28.56 & 6.89 \\
\bottomrule
\end{tabular}
}
\caption{Tagging accuracy of in-vocabulary (non-OOV) and out-of-vocabulary (OOV) words in UDA + 500K in-domain data.}
\label{tab:oov}
\end{center}
\end{table}

\paragraph{Syntactic Diversity in Masked Subset.} Table~\ref{tab:pos_ratio} describes the percentage of POS tags in masked subset selected by different masking strategies. We observed that our method selects more tokens from the major POS tags (71\%) compared to random (45\%) and entropy-based (55\%) strategies. It has chosen less nouns compared to the POS strategy, and more pronouns compared to all other strategies.

\paragraph{Tagging Accuracy of OOV and non-OOV.} We compare the tagging accuracy of out-of-vocabulary (OOV) words which are in target domain but not presenting in source, and non-OOV tokens in Table~\ref{tab:oov}. As seen, our adversarial masking strategy  achieves higher accuracy on both OOV and non-OOV tokens in most cases.


\section{Related Work}
\paragraph{Unsupervised Domain Adaptation.} 

The main approaches in neural UDA include discrepancy-based and adversarial-based methods. The discrepancy-based methods are based on the usage of the maximum mean discrepancy or Wasserstein distance as a regularizer to enforce the learning of domain non-discriminative representations~\citep{Shen2018}. Inspired by the Generative Adversarial Network (GAN)~\citep{Goodfellow2014}, the adversarial-based methods learn a representation that is discriminative for the target task and indiscriminative to the shift between the  domains~\citep{Ganin2015}.

\paragraph{Domain Adaptation with MLM.} Performance of fine-tuned MLM can deteriorate substantially on the presence of domain mismatch. The most straightforward domain adaptation approach in MLM  is to adapt general contextual embedding to a specific domain~\citep{lee2020biobert, alsentzer-etal-2019-publicly, chakrabarty-etal-2019-imho}, that is to further improve pretrained MLM  by continuing to pretrain language models on related domain or similar tasks~\citep{Gururangan2020}, or via intermediate task which is also referred to as STILTs~\citep{phang2018sentence}. Recent works have proposed two-step adaptive domain adaptation framework which consists of domain tuning and task finetuning~\citep{Ma2019, Xu2019, Wang2019,logeswaran-etal-2019-zero}. They have demonstrated that domain tuning is necessary to adapt MLM with both domain knowledge and task knowledge before finetuning, especially when the labelled data in target task is extremely small. Our experiment setting is similar to~\citet{han-eisenstein-2019-unsupervised}'s work. However, we focus on learning masking strategy to boost the domain-tuning step.

\paragraph{Adversarial Learning.} Recent research in adversarial machine learning has either focused on attacking models with adversarial examples~\citep{alzantot-etal-2018-generating, iyyer-etal-2018-adversarial,ebrahimi-etal-2018-hotflip}, or training models to be robust against these attacks~\citep{zhou-etal-2019-learning}.
\citet{pmlr-v97-wang19f,liu2020adversarial} propose the use of adversarial learning for language models. They consider autoregressive LMs and train them to be robust against adversarial perturbations of the word embeddings of the target vocabulary. 


\section{Conclusion}
We present an adversarial objective for further pretraining MLM in UDA problem. The intuition behind the objective is that the adaptation effort should focus on a subset of tokens which are challenging to the MLM. We establish a variational lower bound of the objective function and propose an effective sampling algorithm using dynamic programming and Gumbel softmax trick. Comparing to other masking strategies, our proposed adversarial masking approach has achieve substantially better performance on UDA problem of named entity span prediction for several domains. 

\section*{Acknowledgments}
This material is based on research sponsored by Air Force Research Laboratory and DARPA under agreement number FA8750-19-2-0501. The U.S. Government is authorized to reproduce and distribute reprints for Governmental purposes notwithstanding any copyright notation thereon. The  authors  are  grateful  to  the  anonymous  reviewers for their helpful comments. The computational resources of this work are supported by the Google Cloud Platform (GCP), and by the Multi-modal Australian ScienceS Imaging  and  Visualisation  Environment (MASSIVE) (\url{www.massive.org.au}).

\bibliography{anthology,emnlp2020}

\begin{thebibliography}{39}
\expandafter\ifx\csname natexlab\endcsname\relax\def\natexlab#1{#1}\fi

\bibitem[{Abadi et~al.(2016)Abadi, Barham, Chen, Chen, Davis, Dean, Devin,
  Ghemawat, Irving, Isard et~al.}]{abadi2016tensorflow}
Mart{\'\i}n Abadi, Paul Barham, Jianmin Chen, Zhifeng Chen, Andy Davis, Jeffrey
  Dean, Matthieu Devin, Sanjay Ghemawat, Geoffrey Irving, Michael Isard, et~al.
  2016.
\newblock Tensorflow: A system for large-scale machine learning.
\newblock In \emph{12th $\{$USENIX$\}$ Symposium on Operating Systems Design
  and Implementation ($\{$OSDI$\}$ 16)}, pages 265--283.

\bibitem[{Alsentzer et~al.(2019)Alsentzer, Murphy, Boag, Weng, Jindi, Naumann,
  and McDermott}]{alsentzer-etal-2019-publicly}
Emily Alsentzer, John Murphy, William Boag, Wei-Hung Weng, Di~Jindi, Tristan
  Naumann, and Matthew McDermott. 2019.
\newblock \href {https://doi.org/10.18653/v1/W19-1909} {Publicly available
  clinical {BERT} embeddings}.
\newblock In \emph{Proceedings of the 2nd Clinical Natural Language Processing
  Workshop}, pages 72--78, Minneapolis, Minnesota, USA. Association for
  Computational Linguistics.

\bibitem[{Alzantot et~al.(2018)Alzantot, Sharma, Elgohary, Ho, Srivastava, and
  Chang}]{alzantot-etal-2018-generating}
Moustafa Alzantot, Yash Sharma, Ahmed Elgohary, Bo-Jhang Ho, Mani Srivastava,
  and Kai-Wei Chang. 2018.
\newblock \href {https://doi.org/10.18653/v1/D18-1316} {Generating natural
  language adversarial examples}.
\newblock In \emph{Proceedings of the 2018 Conference on Empirical Methods in
  Natural Language Processing}, pages 2890--2896, Brussels, Belgium.
  Association for Computational Linguistics.

\bibitem[{Brown et~al.(2020)Brown, Mann, Ryder, Subbiah, Kaplan, Dhariwal,
  Neelakantan, Shyam, Sastry, Askell, Agarwal, Herbert-Voss, Krueger, Henighan,
  Child, Ramesh, Ziegler, Wu, Winter, Hesse, Chen, Sigler, Litwin, Gray, Chess,
  Clark, Berner, McCandlish, Radford, Sutskever, and
  Amodei}]{brown2020language}
Tom~B. Brown, Benjamin Mann, Nick Ryder, Melanie Subbiah, Jared Kaplan,
  Prafulla Dhariwal, Arvind Neelakantan, Pranav Shyam, Girish Sastry, Amanda
  Askell, Sandhini Agarwal, Ariel Herbert-Voss, Gretchen Krueger, Tom Henighan,
  Rewon Child, Aditya Ramesh, Daniel~M. Ziegler, Jeffrey Wu, Clemens Winter,
  Christopher Hesse, Mark Chen, Eric Sigler, Mateusz Litwin, Scott Gray,
  Benjamin Chess, Jack Clark, Christopher Berner, Sam McCandlish, Alec Radford,
  Ilya Sutskever, and Dario Amodei. 2020.
\newblock \href {http://arxiv.org/abs/2005.14165} {Language models are few-shot
  learners}.

\bibitem[{Chakrabarty et~al.(2019)Chakrabarty, Hidey, and
  McKeown}]{chakrabarty-etal-2019-imho}
Tuhin Chakrabarty, Christopher Hidey, and Kathy McKeown. 2019.
\newblock \href {https://doi.org/10.18653/v1/N19-1054} {{IMHO} fine-tuning
  improves claim detection}.
\newblock In \emph{Proceedings of the 2019 Conference of the North {A}merican
  Chapter of the Association for Computational Linguistics: Human Language
  Technologies, Volume 1 (Long and Short Papers)}, pages 558--563, Minneapolis,
  Minnesota. Association for Computational Linguistics.

\bibitem[{Clark et~al.(2020)Clark, Luong, Le, and Manning}]{Clark2020}
Kevin Clark, Minh-Thang Luong, Quoc~V. Le, and Christopher~D. Manning. 2020.
\newblock \href {https://openreview.net/forum?id=r1xMH1BtvB} {{\{}ELECTRA{\}}:
  Pre-training text encoders as discriminators rather than generators}.
\newblock In \emph{International Conference on Learning Representations}.

\bibitem[{Collier and Kim(2004)}]{collier-kim-2004-introduction}
Nigel Collier and Jin-Dong Kim. 2004.
\newblock \href {https://www.aclweb.org/anthology/W04-1213} {Introduction to
  the bio-entity recognition task at {JNLPBA}}.
\newblock In \emph{Proceedings of the International Joint Workshop on Natural
  Language Processing in Biomedicine and its Applications
  ({NLPBA}/{B}io{NLP})}, pages 73--78, Geneva, Switzerland. COLING.

\bibitem[{DeSola et~al.(2019)DeSola, Hanna, and Nonis}]{desolafinbert}
Vinicio DeSola, Kevin Hanna, and Pri Nonis. 2019.
\newblock Finbert: pre-trained model on sec filings for financial natural
  language tasks.

\bibitem[{Devlin et~al.(2019)Devlin, Chang, Lee, and
  Toutanova}]{devlin-etal-2019-bert}
Jacob Devlin, Ming-Wei Chang, Kenton Lee, and Kristina Toutanova. 2019.
\newblock \href {https://doi.org/10.18653/v1/N19-1423} {{BERT}: Pre-training of
  deep bidirectional transformers for language understanding}.
\newblock In \emph{Proceedings of the 2019 Conference of the North {A}merican
  Chapter of the Association for Computational Linguistics: Human Language
  Technologies, Volume 1 (Long and Short Papers)}, pages 4171--4186,
  Minneapolis, Minnesota. Association for Computational Linguistics.

\bibitem[{Ebrahimi et~al.(2018)Ebrahimi, Rao, Lowd, and
  Dou}]{ebrahimi-etal-2018-hotflip}
Javid Ebrahimi, Anyi Rao, Daniel Lowd, and Dejing Dou. 2018.
\newblock \href {https://doi.org/10.18653/v1/P18-2006} {{H}ot{F}lip: White-box
  adversarial examples for text classification}.
\newblock In \emph{Proceedings of the 56th Annual Meeting of the Association
  for Computational Linguistics (Volume 2: Short Papers)}, pages 31--36,
  Melbourne, Australia. Association for Computational Linguistics.

\bibitem[{Ganin and Lempitsky(2015)}]{Ganin2015}
Yaroslav Ganin and Victor Lempitsky. 2015.
\newblock Unsupervised domain adaptation by backpropagation.
\newblock In \emph{Proceedings of the 32nd International Conference on
  International Conference on Machine Learning - Volume 37}, ICML’15, page
  1180–1189. JMLR.org.

\bibitem[{Go et~al.(2009)Go, Bhayani, and Huang}]{go2009twitter}
Alec Go, Richa Bhayani, and Lei Huang. 2009.
\newblock Twitter sentiment classification using distant supervision.
\newblock \emph{CS224N project report, Stanford}, 1(12):2009.

\bibitem[{Goodfellow et~al.(2014)Goodfellow, Pouget-Abadie, Mirza, Xu,
  Warde-Farley, Ozair, Courville, and Bengio}]{Goodfellow2014}
Ian Goodfellow, Jean Pouget-Abadie, Mehdi Mirza, Bing Xu, David Warde-Farley,
  Sherjil Ozair, Aaron Courville, and Yoshua Bengio. 2014.
\newblock Generative adversarial nets.
\newblock In \emph{Advances in neural information processing systems}, pages
  2672--2680.

\bibitem[{Gururangan et~al.(2020)Gururangan, Marasović, Swayamdipta, Lo,
  Beltagy, Downey, and Smith}]{Gururangan2020}
Suchin Gururangan, Ana Marasović, Swabha Swayamdipta, Kyle Lo, Iz~Beltagy,
  Doug Downey, and Noah~A. Smith. 2020.
\newblock Don't stop pretraining: Adapt language models to domains and tasks.
\newblock In \emph{Proceedings of ACL}.

\bibitem[{Han and Eisenstein(2019)}]{han-eisenstein-2019-unsupervised}
Xiaochuang Han and Jacob Eisenstein. 2019.
\newblock \href {https://doi.org/10.18653/v1/D19-1433} {Unsupervised domain
  adaptation of contextualized embeddings for sequence labeling}.
\newblock In \emph{Proceedings of the 2019 Conference on Empirical Methods in
  Natural Language Processing and the 9th International Joint Conference on
  Natural Language Processing (EMNLP-IJCNLP)}, pages 4238--4248, Hong Kong,
  China. Association for Computational Linguistics.

\bibitem[{Iyyer et~al.(2018)Iyyer, Wieting, Gimpel, and
  Zettlemoyer}]{iyyer-etal-2018-adversarial}
Mohit Iyyer, John Wieting, Kevin Gimpel, and Luke Zettlemoyer. 2018.
\newblock \href {https://doi.org/10.18653/v1/N18-1170} {Adversarial example
  generation with syntactically controlled paraphrase networks}.
\newblock In \emph{Proceedings of the 2018 Conference of the North {A}merican
  Chapter of the Association for Computational Linguistics: Human Language
  Technologies, Volume 1 (Long Papers)}, pages 1875--1885, New Orleans,
  Louisiana. Association for Computational Linguistics.

\bibitem[{Jang et~al.(2017)Jang, Gu, and Poole}]{gumbel}
Eric Jang, Shixiang Gu, and Ben Poole. 2017.
\newblock Categorical reparameterization with gumbel-softmax.
\newblock In \emph{International Conference on Learning Representations}.

\bibitem[{Kim et~al.(2009)Kim, Ohta, Pyysalo, Kano, and
  Tsujii}]{kim-etal-2009-overview}
Jin-Dong Kim, Tomoko Ohta, Sampo Pyysalo, Yoshinobu Kano, and Jun{'}ichi
  Tsujii. 2009.
\newblock \href {https://www.aclweb.org/anthology/W09-1401} {Overview of
  {B}io{NLP}{'}09 shared task on event extraction}.
\newblock In \emph{Proceedings of the {B}io{NLP} 2009 Workshop Companion Volume
  for Shared Task}, pages 1--9, Boulder, Colorado. Association for
  Computational Linguistics.

\bibitem[{Kim et~al.(2011)Kim, Pyysalo, Ohta, Bossy, Nguyen, and
  Tsujii}]{kim-etal-2011-overview}
Jin-Dong Kim, Sampo Pyysalo, Tomoko Ohta, Robert Bossy, Ngan Nguyen, and
  Jun{'}ichi Tsujii. 2011.
\newblock \href {https://www.aclweb.org/anthology/W11-1801} {Overview of
  {B}io{NLP} shared task 2011}.
\newblock In \emph{Proceedings of {B}io{NLP} Shared Task 2011 Workshop}, pages
  1--6, Portland, Oregon, USA. Association for Computational Linguistics.

\bibitem[{Lee et~al.(2020)Lee, Yoon, Kim, Kim, Kim, So, and
  Kang}]{lee2020biobert}
Jinhyuk Lee, Wonjin Yoon, Sungdong Kim, Donghyeon Kim, Sunkyu Kim, Chan~Ho So,
  and Jaewoo Kang. 2020.
\newblock Biobert: a pre-trained biomedical language representation model for
  biomedical text mining.
\newblock \emph{Bioinformatics}, 36(4):1234--1240.

\bibitem[{Liu et~al.(2020)Liu, Cheng, He, Chen, Wang, Poon, and
  Gao}]{liu2020adversarial}
Xiaodong Liu, Hao Cheng, Pengcheng He, Weizhu Chen, Yu~Wang, Hoifung Poon, and
  Jianfeng Gao. 2020.
\newblock Adversarial training for large neural language models.
\newblock \emph{arXiv preprint arXiv:2004.08994}.

\bibitem[{Liu et~al.(2019)Liu, Ott, Goyal, Du, Joshi, Chen, Levy, Lewis,
  Zettlemoyer, and Stoyanov}]{roberta}
Yinhan Liu, Myle Ott, Naman Goyal, Jingfei Du, Mandar Joshi, Danqi Chen, Omer
  Levy, Mike Lewis, Luke Zettlemoyer, and Veselin Stoyanov. 2019.
\newblock \href {http://arxiv.org/abs/1907.11692} {Roberta: {A} robustly
  optimized {BERT} pretraining approach}.
\newblock \emph{CoRR}, abs/1907.11692.

\bibitem[{Logeswaran et~al.(2019)Logeswaran, Chang, Lee, Toutanova, Devlin, and
  Lee}]{logeswaran-etal-2019-zero}
Lajanugen Logeswaran, Ming-Wei Chang, Kenton Lee, Kristina Toutanova, Jacob
  Devlin, and Honglak Lee. 2019.
\newblock \href {https://doi.org/10.18653/v1/P19-1335} {Zero-shot entity
  linking by reading entity descriptions}.
\newblock In \emph{Proceedings of the 57th Annual Meeting of the Association
  for Computational Linguistics}, pages 3449--3460, Florence, Italy.
  Association for Computational Linguistics.

\bibitem[{Ma et~al.(2019)Ma, Xu, Wang, Nallapati, and Xiang}]{Ma2019}
Xiaofei Ma, Peng Xu, Zhiguo Wang, Ramesh Nallapati, and Bing Xiang. 2019.
\newblock \href {https://doi.org/10.18653/v1/D19-6109} {Domain adaptation with
  {BERT}-based domain classification and data selection}.
\newblock In \emph{Proceedings of the 2nd Workshop on Deep Learning Approaches
  for Low-Resource NLP (DeepLo 2019)}, pages 76--83, Hong Kong, China.
  Association for Computational Linguistics.

\bibitem[{Moore and Lewis(2010)}]{moore-lewis-2010-intelligent}
Robert~C. Moore and William Lewis. 2010.
\newblock \href {https://www.aclweb.org/anthology/P10-2041} {Intelligent
  selection of language model training data}.
\newblock In \emph{Proceedings of the {ACL} 2010 Conference Short Papers},
  pages 220--224, Uppsala, Sweden. Association for Computational Linguistics.

\bibitem[{Peters et~al.(2018)Peters, Neumann, Iyyer, Gardner, Clark, Lee, and
  Zettlemoyer}]{peters-etal-2018-deep}
Matthew Peters, Mark Neumann, Mohit Iyyer, Matt Gardner, Christopher Clark,
  Kenton Lee, and Luke Zettlemoyer. 2018.
\newblock \href {https://doi.org/10.18653/v1/N18-1202} {Deep contextualized
  word representations}.
\newblock In \emph{Proceedings of the 2018 Conference of the North {A}merican
  Chapter of the Association for Computational Linguistics: Human Language
  Technologies, Volume 1 (Long Papers)}, pages 2227--2237, New Orleans,
  Louisiana. Association for Computational Linguistics.

\bibitem[{Phang et~al.(2018)Phang, F{\'e}vry, and Bowman}]{phang2018sentence}
Jason Phang, Thibault F{\'e}vry, and Samuel~R Bowman. 2018.
\newblock Sentence encoders on stilts: Supplementary training on intermediate
  labeled-data tasks.
\newblock \emph{arXiv preprint arXiv:1811.01088}.

\bibitem[{Raffel et~al.(2019)Raffel, Shazeer, Roberts, Lee, Narang, Matena,
  Zhou, Li, and Liu}]{2019t5}
Colin Raffel, Noam Shazeer, Adam Roberts, Katherine Lee, Sharan Narang, Michael
  Matena, Yanqi Zhou, Wei Li, and Peter~J. Liu. 2019.
\newblock \href {http://arxiv.org/abs/1910.10683} {Exploring the limits of
  transfer learning with a unified text-to-text transformer}.
\newblock \emph{arXiv e-prints}.

\bibitem[{Ruder and Plank(2017)}]{ruder-plank-2017-learning}
Sebastian Ruder and Barbara Plank. 2017.
\newblock \href {https://doi.org/10.18653/v1/D17-1038} {Learning to select data
  for transfer learning with {B}ayesian optimization}.
\newblock In \emph{Proceedings of the 2017 Conference on Empirical Methods in
  Natural Language Processing}, pages 372--382, Copenhagen, Denmark.
  Association for Computational Linguistics.

\bibitem[{Salinas~Alvarado et~al.(2015)Salinas~Alvarado, Verspoor, and
  Baldwin}]{salinas-alvarado-etal-2015-domain}
Julio~Cesar Salinas~Alvarado, Karin Verspoor, and Timothy Baldwin. 2015.
\newblock \href {https://www.aclweb.org/anthology/U15-1010} {Domain adaption of
  named entity recognition to support credit risk assessment}.
\newblock In \emph{Proceedings of the Australasian Language Technology
  Association Workshop 2015}, pages 84--90, Parramatta, Australia.

\bibitem[{Shen et~al.(2018)Shen, Qu, Zhang, and Yu}]{Shen2018}
Jian Shen, Yanru Qu, Weinan Zhang, and Yong Yu. 2018.
\newblock Wasserstein distance guided representation learning for domain
  adaptation.
\newblock In \emph{Thirty-Second AAAI Conference on Artificial Intelligence}.

\bibitem[{Smith et~al.(2008)Smith, Tanabe, nee Ando, Kuo, Chung, Hsu, Lin,
  Klinger, Friedrich, Ganchev et~al.}]{smith2008overview}
Larry Smith, Lorraine~K Tanabe, Rie~Johnson nee Ando, Cheng-Ju Kuo, I-Fang
  Chung, Chun-Nan Hsu, Yu-Shi Lin, Roman Klinger, Christoph~M Friedrich, Kuzman
  Ganchev, et~al. 2008.
\newblock Overview of biocreative ii gene mention recognition.
\newblock \emph{Genome biology}, 9(2):S2.

\bibitem[{Strauss et~al.(2016)Strauss, Toma, Ritter, de~Marneffe, and
  Xu}]{strauss-etal-2016-results}
Benjamin Strauss, Bethany Toma, Alan Ritter, Marie-Catherine de~Marneffe, and
  Wei Xu. 2016.
\newblock \href {https://www.aclweb.org/anthology/W16-3919} {Results of the
  {WNUT}16 named entity recognition shared task}.
\newblock In \emph{Proceedings of the 2nd Workshop on Noisy User-generated Text
  ({WNUT})}, pages 138--144, Osaka, Japan. The COLING 2016 Organizing
  Committee.

\bibitem[{Wang et~al.(2019{\natexlab{a}})Wang, Singh, Michael, Hill, Levy, and
  Bowman}]{wang2019glue}
Alex Wang, Amanpreet Singh, Julian Michael, Felix Hill, Omer Levy, and
  Samuel~R. Bowman. 2019{\natexlab{a}}.
\newblock {GLUE}: A multi-task benchmark and analysis platform for natural
  language understanding.
\newblock In the Proceedings of ICLR.

\bibitem[{Wang et~al.(2019{\natexlab{b}})Wang, Gong, and
  Liu}]{pmlr-v97-wang19f}
Dilin Wang, Chengyue Gong, and Qiang Liu. 2019{\natexlab{b}}.
\newblock Improving neural language modeling via adversarial training.
\newblock In \emph{Proceedings of the 36th International Conference on Machine
  Learning}, volume~97 of \emph{Proceedings of Machine Learning Research},
  pages 6555--6565, Long Beach, California, USA.

\bibitem[{Wang et~al.(2019{\natexlab{c}})Wang, Gan, Liu, Liu, Gao, and
  Wang}]{Wang2019}
Huazheng Wang, Zhe Gan, Xiaodong Liu, Jingjing Liu, Jianfeng Gao, and Hongning
  Wang. 2019{\natexlab{c}}.
\newblock \href {https://doi.org/10.18653/v1/D19-1254} {Adversarial domain
  adaptation for machine reading comprehension}.
\newblock In \emph{Proceedings of the 2019 Conference on Empirical Methods in
  Natural Language Processing and the 9th International Joint Conference on
  Natural Language Processing (EMNLP-IJCNLP)}, pages 2510--2520, Hong Kong,
  China. Association for Computational Linguistics.

\bibitem[{Xu et~al.(2019)Xu, Liu, Shu, and Yu}]{Xu2019}
Hu~Xu, Bing Liu, Lei Shu, and Philip Yu. 2019.
\newblock \href {https://doi.org/10.18653/v1/N19-1242} {{BERT} post-training
  for review reading comprehension and aspect-based sentiment analysis}.
\newblock In \emph{Proceedings of the 2019 Conference of the North {A}merican
  Chapter of the Association for Computational Linguistics: Human Language
  Technologies, Volume 1 (Long and Short Papers)}, pages 2324--2335,
  Minneapolis, Minnesota. Association for Computational Linguistics.

\bibitem[{Yang et~al.(2019)Yang, Dai, Yang, Carbonell, Salakhutdinov, and
  Le}]{yang2019xlnet}
Zhilin Yang, Zihang Dai, Yiming Yang, Jaime Carbonell, Russ~R Salakhutdinov,
  and Quoc~V Le. 2019.
\newblock Xlnet: Generalized autoregressive pretraining for language
  understanding.
\newblock In \emph{Advances in neural information processing systems}, pages
  5754--5764.

\bibitem[{Zhou et~al.(2019)Zhou, Shah, and
  Schockaert}]{zhou-etal-2019-learning}
Yilun Zhou, Julie Shah, and Steven Schockaert. 2019.
\newblock \href {https://www.aclweb.org/anthology/W19-5808} {Learning household
  task knowledge from {W}iki{H}ow descriptions}.
\newblock In \emph{Proceedings of the 5th Workshop on Semantic Deep Learning
  (SemDeep-5)}, pages 50--56, Macau, China. Association for Computational
  Linguistics.

\end{thebibliography}
\bibliographystyle{acl_natbib}

\end{document}